# Automatic local Gabor features extraction for face recognition

Yousra BEN JEMAA

National Engineering School of Sfax
Signal and System Unit
Tunisia
yousra.benjemaa@planet.tn

Sana KHANFIR

National Engineering School of Sfax
Tunisia

*Abstract*—We present in this paper a biometric system of face detection and recognition in color images. The face detection technique is based on skin color information and fuzzy classification. A new algorithm is proposed in order to detect automatically face features (eyes, mouth and nose) and extract their correspondent geometrical points. These fiducial points are described by sets of wavelet components which are used for recognition. To achieve the face recognition, we use neural networks and we study its performances for different inputs. We compare the two types of features used for recognition: geometric distances and Gabor coefficients which can be used either independently or jointly. This comparison shows that Gabor coefficients are more powerful than geometric distances. We show with experimental results how the importance recognition ratio makes our system an effective tool for automatic face detection and recognition.

I. INTRODUCTION

Face recognition is a very challenging area in computer vision and pattern recognition due to variations in facial expressions, poses, illumination. Face recognition is largely motivated by the need for access control, surveillance and security, telecommunication and digital library...[10].

Face detection is the first stage of an automated face recognition system, since a face has to be located before it is recognized [11]. Consequently, the performance of face detection step certainly affects the performance of the recognition system. The problem of face detection is a very challenging due to the diverse variation of face and the complexity of background in images. Different methods have been proposed, they can be classified into four categories: Knowledge-based methods, Feature-based methods, Template-based methods and Appearance-based methods [12].

We present a face detection technique based on the skin color classification.

When skin regions are detected, facial feature extraction attempts to find the most appropriate representation of the face images for recognition. There are mainly two approaches: General systems and Geometric feature-based systems [10]. In general methods, faces are treated as a whole object. In order to reduce the dimensionality of the face representation, principal component analysis (PCA) or eigenfaces [8] and neural networks are extensively used. Geometrical methods are based on measures extracted between the facial features such as eyes, mouth, nose and chin. Representative works include hidden Markov model (HMM) [19], elastic bunch graph matching algorithm [3] and local feature analysis.

Geometrical techniques are usually computationally more expensive than global techniques but are more robust to variations (size, orientation...).

We use the two techniques; we first locate the feature points and then apply Gabor filters in each point in order to extract a set of Gabor wavelet coefficients. We use wavelet analysis because it can localize, in space-frequency, characteristics of images and it can represent faces in different spatial resolutions and orientations [7] [17]. For example, Zhang [9] has proposed a system of face recognition based on Gabor wavelet associative memory (a neural network)[6]. Wiskott [3] proposed a face recognition system by elastic bunch graph matching on which he represents the local features of faces (eyes, mouth, etc.) by a set of Gabor wavelets.

The proposed system of face recognition can detect the face in an image and localize automatically 10 fiducial points. After that it characterizes the face by the geometric distances between the extracted points or by applying a set of Gabor wavelets (filters) in correspondence to each fiducial point. These two types of features extracted from face can be used either independently or jointly. They are used as input data to neural networks for classification. The recognition performances with different types of features are compared.

The paper is organized as follows: the face detection system is described in section 2. Section 3 presents the algorithm of facial features localization and extraction of their characteristic points. The two types of features extracted from face are described in section 4. The face recognition system based on neural networks is presented in section 5. Section 6 provides the experimental results and the comparison between the two types of features. Finally, we conclude in section 7.

II. THE FACE DETECTION SYSTEM

Face detection is a very important stage to start the step of face recognition. In fact, to identify a person, it is necessary to localize his face in the image. Our system of detection supposes that only one face is presented in the field of the camera. It is described in fig 1.

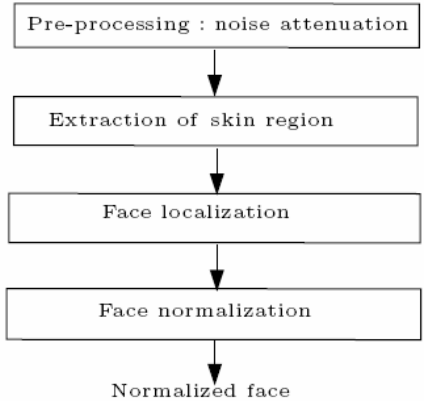

Figure 1. Steps of face detection

*A. Preprocessing and skin color extraction*

In the first step, an average filter, a low-pass filter, is applied to the image to attenuate the noise. In the second step, the methodology of face detection is assured by the skin color technique since it is invariant to changes in size, orientations and occlusion...Many color spaces have been proposed in the literature for skin detection [13]. We propose for our system to use the YCbCr color space, because the Cr and Cb (chrominance) components are independent of the skin color, the human race and the lighting conditions (it is in the YCbCr color space that the luminance is decoupled from the color information). Furthermore, the Cb and Cr are the chrominance components used in MPEG and JPEG [20].

Fig 2 illustrates how the two chrominance plans have separated the skin color from the background.

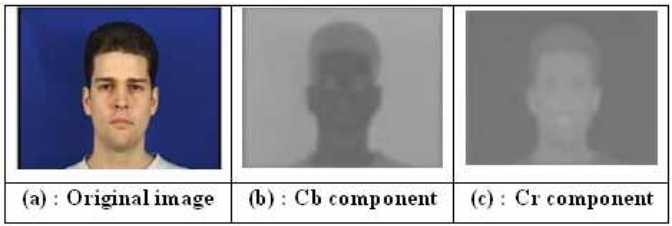

Figure 2. Conversion of an RGB image (a) into Cb (b) and Cr (c) space

In order to classify each pixel into skin or non-skin pixel, the most suitable arrangements that we found for all input images in database are [4]: Cb in [77, 127] and Cr in [133, 173]. These arrangements are not sufficient to find a good classification. Fig 3 shows the limitation of this approach for classification. In fact, a big part of the face is considered as background.

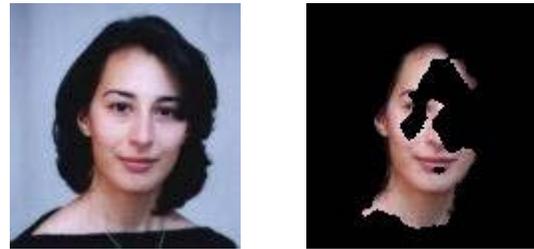

Figure 3. Classification of the image into skin region and non-skin region

To overcome the problem listed above, we propose to apply a fuzzy approach for pixel classification. This is considered as a good solution since that, fuzzy set theory can represent and manipulate uncertainly and ambiguity [15]. We use the Takagi-Sugeno fuzzy inference system (FIS). This system is composed of two inputs (the 2 components Cb and Cr) and one output (the decision: skin or non-skin color). Each input has three sub-sets: light, medium and dark. Our algorithm uses the concept of fuzzy logic IF-THEN rules; these rules are applied in each pixel in the image in order to decide whether the pixel represents a skin or non-skin region [21].

The fuzzy logic rules applied for skin detection are the following:
1. IF Cb is Light and Cr is Light THEN the pixel =0
2. IF Cb is Light and Cr is Medium THEN the pixel =0
3. IF Cb is Light and Cr is Dark THEN the pixel =0
4. IF Cb is Medium and Cr is Light THEN the pixel =0
5. IF Cb is Medium and Cr is Medium THEN the pixel =1
6. IF Cb is Medium and Cr is Dark THEN the pixel =1
7. IF Cb is Dark and Cr is Light THEN the pixel =0
8. IF Cb is Dark and Cr is Medium THEN the pixel =1
9. IF Cb is Dark and Cr is Dark THEN the pixel =0

The first step is to determine, for each input, the degree of membership to the appropriate fuzzy sets via membership functions. Once the inputs have been fuzzified, the final decision of the inference system is the average of the output ($z_i$) corresponding to the rule ($r_i$) weighted by the normalized degree $p_i$ of the rule (1 or 0).

Fig 4 represents the input image and the output one after YCbCr color space conversion and skin region extraction using fuzzy classification.

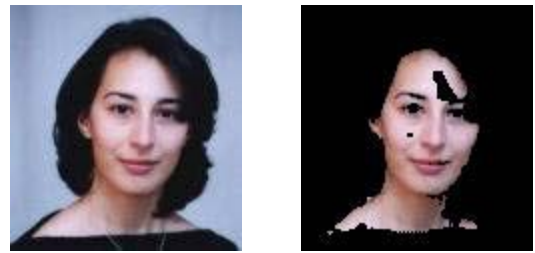

Figure 4. Classification of the image into skin region and non-skin region using the YCbCr space and fuzzy classification

A comparison of figures 3 and 4 proves the importance of fuzzy logic in skin detection. We note that the number of the skin pixels detected by fuzzy logic is more important than the number using classic classification. Skin regions are represented in white and Non skin regions are represented in black. This step allows detecting skin regions in images but it is not sufficient to detect the external edge of the face.

*B. Face localization and normalization*

Canny edge detection [16] is selected in this process to detect the external edge of the face. The method differs from the other edge detection methods in that it includes the weak edges in the output only if they are connected to the strong edges. After that, we extract the extreme points of this edge in the horizontal sense and the highest point of the face. The lowest point of the face is determined while supposing that the height of the face is bigger than the width about 1.3 times. Finally, the rectangle that delimits the face is defined by these four points as shown in fig 5. A standard size for all face examples of the database is necessary to lead a relatively correct recognition operation. In this work, images have a standard size of 50×50.

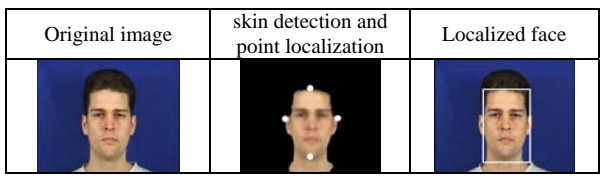

Figure 5. Face detection steps

The following step consists of extracting face vector characteristic from normalized face.

### III. AUTOMATIC FACIAL FEATURE LOCALIZATION AND EXTRACTION OF THE FACE CHARACTERISTIC POINTS

The facial features used in our system of face recognition are: eyes, mouth and nose. The system is described in fig 6.

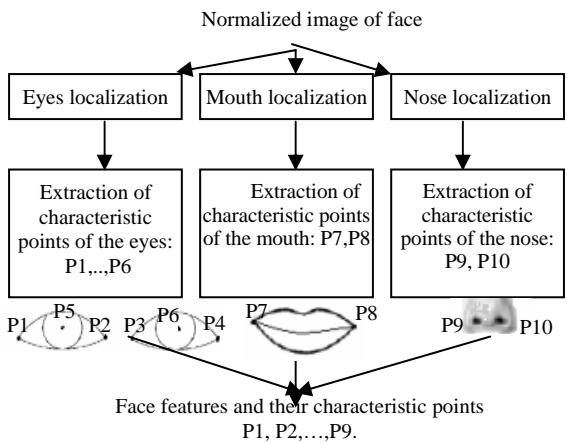

Figure 6. System of features localization

Analysis of the chrominance components indicates that eyes present high values of Cb and small values of Cr [5]. So, we first extract the chrominance component Cb and then we find its maximal value. We proceed then to a threshold in order to obtain a black and white image. The zone of the eyes is divided in two equal parts, after that we find in every zone the white stains of maximal area. Finally, we apply an operation of dilation by a flat disk-shaped structuring element. We extract four points representing the external points of each maximal area of stains representing each eye. We also determine the center of gravity of these white stains that represents the center of each eye. Finally, we obtain 6 points characterizing the two eyes.

The region of the mouth is in the bottom part of the face. Since the component of chrominance Cr depends on the red color [5], we use it to localize the mouth. In order to localize its geometrical points, we apply a Sobel filter to detect its contour then we extract its extreme points [2].

After the detection of the mouth and eyes, the localization of the nose becomes obvious since it is located between them. The geometrical points are also extracted by using a Sobel filter [2].

Fig 7 shows the experimental results when applying our system of features localization.

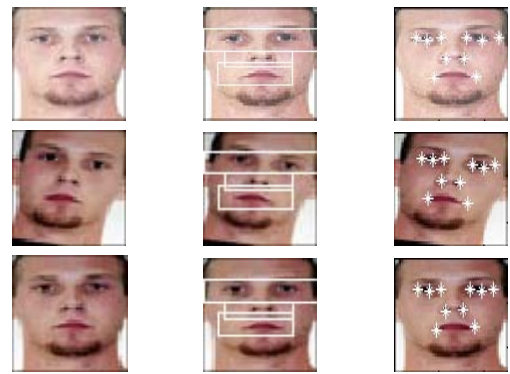

Figure 7. Facial features localization and extraction of the characteristic points

### IV. FACIAL FEATURES

*A. Geometric distances*

The elements of the facial feature vector represent significant distances between all extracted points. As shown in fig 8, these distances are:

- *Deye*: The mean of the two distances P1P2 and P3P4.
- *Dcenter_ eye*: The distance between the centers of the two eyes: P5P6.
- *Dinterior_ eye*: Distance between the two eyes: P2P3.
- *Dnose*: The width of the nose: P9P10.
- *Deye_nose*: The height of the nose.
- *Dmouth*: The width of the mouth: P7P8.
- *Dnose_mouth*: The distance between the mouth and the nose.

The facial feature vector is then:

V = [Dcenter_eye; Deye; Dinterior_eye; Dnose; Deye_nose; Dmouth; Dnose_ mouth]

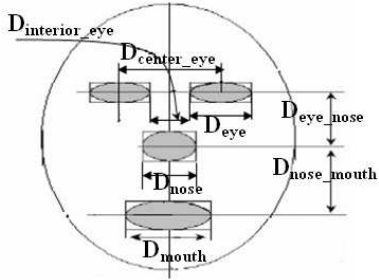

Figure 8. Components of facial features vector

This vector represents each person by a unique way, so it will be used for the recognition step.

### B. Gabor features

Since features extraction for face recognition using Gabor filters is reported to yield good results [10], we use here a Gabor wavelets based feature extraction technique. We use the following family of two-dimensional Gabor kernels [1] [14]:

$$\begin{cases} W(x,y,\theta,\lambda,\varphi,\sigma,\gamma) = \exp(-\frac{x'^2+\gamma^2 y'^2}{2\sigma^2})\cos(2\Pi\frac{x'}{\lambda}+\varphi) \\ x' = x\cos(\theta) + y\sin(\theta) \\ y' = -x\sin(\theta) + y\cos(\theta) \end{cases} \quad (1)$$

where ($x$, $y$) specify the position of a light impulse in the visual field and $\mu$, $\varphi$, $\gamma$, $\lambda$, $\sigma$ are parameters of the wavelet. We have chosen the same parameters used by Wiskott [3] (TABLE I).

TABLE I. PARAMETERS OF GABOR WAVELETS

| Parameter | Symbol | Values |
|---|---|---|
| Orientation | $\theta$ | $\{0, \frac{\Pi}{8}, \frac{2\Pi}{8}, \frac{3\Pi}{8}, \frac{4\Pi}{8}, \frac{5\Pi}{8}, \frac{6\Pi}{8}, \frac{7\Pi}{8}\}$ |
| Wavelength | $\lambda$ | $\{4, 4\sqrt{2}, 8, 8\sqrt{2}, 16\}$ |
| Phase | $\varphi$ | $\{0, \frac{\Pi}{2}\}$ |
| Gaussian Radius | $\sigma$ | $\sigma = \lambda$ |
| Aspect Ratio | $\gamma$ | 1 |

A set of Gabor filters is used with 5 spatial frequencies and 8 distinct orientations, this makes 40 different Gabor filters represented in fig 9.

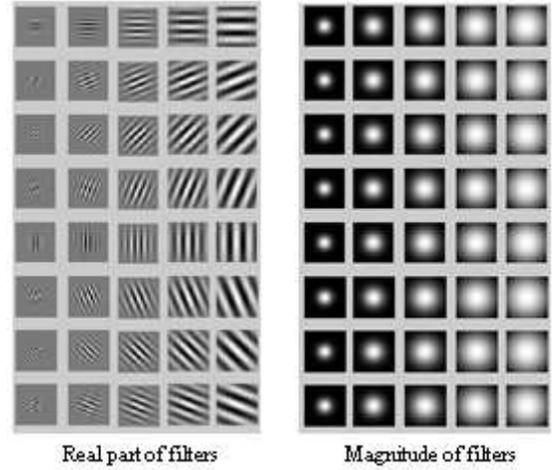

Figure 9. The Gabor filters [3]

When convolving these Gabor filters with a simple face image, we obtain the filter responses. We find out that these representations display desirable locality and orientation performance.

We have selected 5 sets of Gabor filters with different orientations (TABLE II).

TABLE II. SETS OF GABOR FILTERS FOR DIFFERENT ORIENTATIONS

| The number of Gabor filters | The orientations |
|---|---|
| 5 | $\theta = \{0\}$ |
| 10 | $\theta = \{0, \frac{\Pi}{8}, \frac{7\Pi}{8},\}$ |
| 15 | $\theta = \{0, \frac{2\Pi}{8}, \frac{4\Pi}{8}\}$ |
| 20 | $\theta = \{0, \frac{2\Pi}{8}, \frac{4\Pi}{8}, \frac{6\Pi}{8},\}$ |
| 25 | $\theta = \{0, \frac{\Pi}{8}, \frac{2\Pi}{8}, \frac{4\Pi}{8}, \frac{6\Pi}{8},\}$ |

When Gabor filters are applied to each pixel of the image, the dimension of the filtered vector can be very large (proportional to the image dimension). So, it will lead to expensive computation and storage cost. To alleviate such problem and make the algorithm robust, Gabor features are obtained only at the ten extracted fiducial points. If we note C the number of Gabor filters, each point will be represented by a vector of C components called "Jet". When applying the Gabor filters to all fiducial points, we obtain a jets vector of 10× C real coefficients characterizing the face.

### V. FACE RECOGNITION

The face recognition was assured by a non linear classifier which is the neural networks. The advantage using neural networks for recognition is the feasibility of training a system in very complex conditions (rotation, lighting). However, the network architecture has to be varied (number of layers and nodes) to get good performances.

We tried to use a perceptron multi-layer architecture [18]. In order to describe the proposed architecture, we consider P

different persons so *P* classes: there are as many classes as person to identify.

For each person, we have different samples obtained by rotation, by translation and by variation of the lighting. The Neural network is composed by a set of neural networks. There are as many networks as person to identify (P is the number of classes in the database). Each neural network is composed by N input nodes corresponding to the N extracted parameters and one output node which can be active (output=1) or inactive (output=0).

The number of extracted parameters N is fixed when choosing the types of face features. In our study, we have used three types of features :

- Geometric distances between fiducial points.
- Gabor coefficients.
- Combined information between Gabor coefficients and geometric distances.

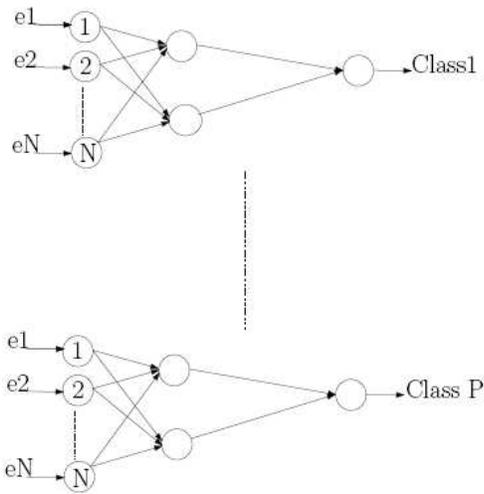

Figure 10. Architecture of the proposed neural network

## VI. EXPERIMENTAL RESULTS

### A. Face database

To test the performances of our recognition system, we have carried out an experiment on the FERET database [22].
The database is split into two subsets at random:
- 60% of the database was used for training and 40% were acquired for testing process,
- 50% of the database was used for training and 50% were acquired for testing process,
- 30% of the database was used for training and 70% were acquired for testing process.

Each of these percentages has been tested on five random combinations of face samples.

### B. Performabces of our method :comparison of different features

Experimental results of varying the system parameters are shown in tables III-V. Particularly, we give the average recognition rate of our system using respectively: geometric distances, Gabor coefficients and the fusion of the two features.

TABLE III. AVERAGE RECOGNITION RATE USING GEOMETRIC DISTANCES

| Training/Test | NN |
|---|---|
| 60%-40% | 84.2105 % |
| 50%-50% | 82.5694 % |
| 30%-70% | 79.7980 % |

TABLE IV. AVERAGE RECOGNITION RATE USING THE GABOR COEFFICIENTS

|  | 60%-40% | 50%-50% | 30%-70% |
|---|---|---|---|
| 5 Gabor wavelets | 98.7719 % | 98.8889 % | 95.3535 % |
| 10 Gabor wavelets | 99.2105 % | 98.8194 % | 98.9394 % |
| 15 Gabor wavelets | 98.5088 % | 99.3056 % | 99.1414 % |
| 20 Gabor wavelets | 99.5614 % | 99.5139 % | 99.2424 % |
| 25 Gabor wavelets | 99.4737 % | 99.5833 % | 98.8889 % |

TABLE V. AVERAGE RECOGNITION RATE USING THE FUSION OF GABOR COEFFICIENTS AND GEOMETRIC DISTANCES

|  | 60%-40% | 50%-50% | 30%-70% |
|---|---|---|---|
| 5 Gabor wavelets | 98.8596 % | 98.1944 % | 96.4646 % |
| 10 Gabor wavelets | 99.2982 % | 99.0278 % | 98.2828 % |
| 15 Gabor wavelets | 99.3860 % | 98.7500 % | 98.5859 % |
| 20 Gabor wavelets | 99.2982 % | 99.3750 % | 99.2424 % |
| 25 Gabor wavelets | 99.6491 % | 99.7222 % | 99.9899 % |

Gabor wavelets can represent features points in special frequency at different orientations. Therefore, the recognition rates, given by the neural network, increase when the number of Gabor wavelets and the number of training set increase. This is shown in fig 11.

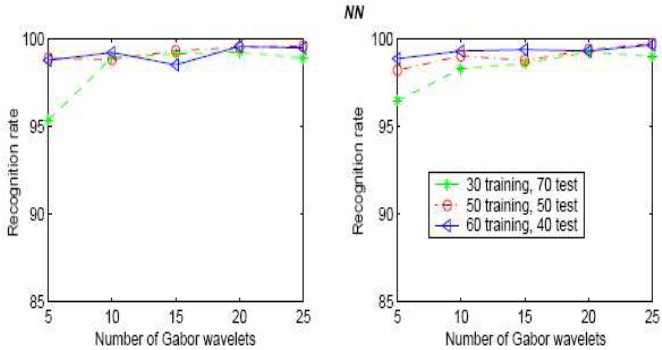

Figure 11. The recognition rate using (1) first row: the Gabor coefficients (2) second row: the Gabor coefficients and geometric distances.

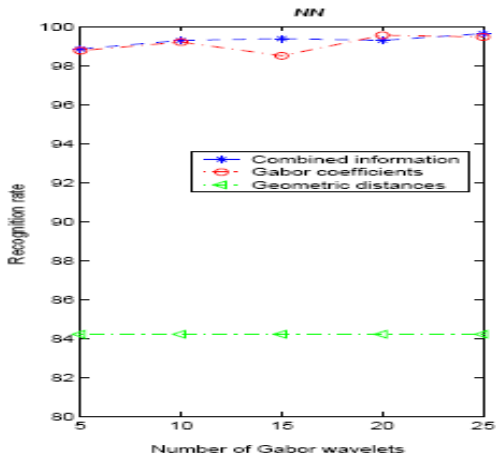

Figure 12. The recognition rate versus Gabor wavelets number for different facial feature vectors

According to fig 12, we deduce that:

- Gabor coefficients are much more powerful than geometric distances.

- Among three types of features, the fusion of Gabor coefficients and geometric distances achieves the highest recognition rate because it takes advantages of the merits of Gabor wavelets and local features.

- More the number of Gabor wavelets increases more the recognition rate increases.

## VII. CONCLUSION

We propose a new system for human face detection and recognition. A contribution of the method is the use of fuzzy classification in order to detect faces. Another contribution is the use of the Gabor wavelets to represent a face since they can represent images in different frequencies at different orientations. Face is represented with its own Gabor coefficients expressed at the fiducial points (points of eyes, mouth and nose).

We have implemented and studied a neural network architecture trained by three characteristic vectors. The first is composed by geometrical distances automatically extracted, by our system, between the fiducial points, the second is composed by the responses of Gabor wavelets applied in the fiducial points and the third is composed by the combined information between the previous vectors. A comparative study between them shows that the third type of features has achieved higher recognition rate (99.98%) for faces at different degrees of rotation and different lighting due to the fusion of Gabor wavelets and local features.

Our future orientation concerns the use of the proposed approach: Gabor wavelets applied in the fiducial points, on 3D faces in order to take into account more features for best recognition.